\documentclass[runningheads]{llncs}
\usepackage{graphicx}
\usepackage{url}
\usepackage{mwe}
\begin{document}
\title{Deep neural network or dermatologist?}
%
%
\author{Kyle Young\inst{1} \and
Gareth Booth\inst{1}\and
Becks Simpson\inst{2}\and 
Reuben Dutton\inst{1}\and
Sally Shrapnel\inst{1}}
\authorrunning{K.Young et al.}
%
\institute{School of Mathematics and Physics, University of Queensland, Brisbane, Australia
\email{s.shrapnel@uq.edu.au}\\
\and
Montreal Institute for Learning Algorithms, Canada}

\maketitle              
\begin{abstract}

Deep learning techniques have proven high accuracy for identifying melanoma in digitised dermoscopic images. A strength is that these methods are not constrained by features that are pre-defined by human semantics. A down-side is that it is difficult to understand the rationale of the model predictions and to identify potential failure modes.  This is a major barrier to adoption of deep learning in clinical practice. In this paper we ask if two existing local interpretability methods, Grad-CAM and Kernel SHAP, can shed light on convolutional neural networks trained in the context of melanoma detection. Our contributions are (i) we first explore the domain space via a reproducible, end-to-end learning framework that creates a suite of 30 models, all trained on a publicly available data set (HAM10000), (ii) we next explore the reliability of GradCAM and Kernel SHAP in this context via some basic sanity check experiments (iii) finally, we investigate a random selection of models from our suite using GradCAM and Kernel SHAP. We show that despite high accuracy, the models will occasionally assign importance to features that are not relevant to the diagnostic task. We also show that models of similar accuracy will produce different explanations as measured by these methods. This work represents first steps in bridging the gap between model accuracy and interpretability in the domain of skin cancer classification. 

\keywords{Deep learning  \and Explainability \and melanoma}
\end{abstract}
\section{Introduction}

Skin cancer is the most common form of cancer in the United States~\cite{rogers2010incidence,lomas2012systematic}, and melanoma is the leading cause of skin cancer related death~\cite{schadendorf2018melanoma}. Automated diagnosis of melanoma from digitized dermoscopy images thus represents an important potential use case for deep learning methods. Inspired by a breakthrough result by Esteva et. al.,~\cite{esteva2017dermatologist}, many recent publications claim ``better than dermatologist" performance of convolutional neural networks (CNNs) on a variety of skin cancer classification tasks~\cite{esteva2017dermatologist,review2018,haenssle2018man,brinker2019deep,Mahbod2019,fujisawa2019deep}. If indeed such models have diagnostic performance comparable to board certified dermatologists, this heralds a new era in skin cancer care, with standardization of diagnosis and democratization of access~\cite{janda2019can,mar2018}. Early diagnosis of melanoma is associated with improved outcomes but poor availability of well trained clinicians in many parts of the world means too often diagnosis is made too late. CNNs represent an important new technology to address this problem for social good.

How can we evaluate the veracity of these exciting new claims? Unfortunately, privacy constraints typically make it difficult to access training, validation, test data, and final model weights. This makes it impossible to verify the accuracy of these published models and reproduce their claims~\cite{review2018}. As is common in medical settings, there are inherent \emph{known} biases in the data: lesion classes are unevenly distributed, healthy images are over-represented, racial bias is present (few lesions are from dark-skinned individuals)~\cite{brinker2019deep} and there is significant variability in ground truth labelling~\cite{Elmorej2813}. Can we be confident the model has not inherited any of these known biases? A further challenge is due to the presence of \emph{unknown} biases in the data. If an artifact is present in images from two diagnostic classes but more prevalent in one, how do we know when the model classifications are weighted by the presence or absence of this artifact?

In light of these problems, it is an open question as to the best strategy to determine if a given model will generalize to future data where the distribution of these biases may be different. Currently there are two approaches: (i) ad hoc techniques that penalize model complexity (batch normalization and dropout, for example), and (ii) training and testing models on larger and more complex data sets. Importantly, neither of these techniques can identify, nor correct, specific biases prior to model deployment.  

In this paper, we investigate the possibility that current interpretability methods may assist in this task. Interpretability methods seek to produce an indication of features of the input data that the model regards as important for weighting the final diagnostic decision. While they do not capture the entirety of the predictive process, they can nonetheless provide some guidance to how a given model makes decisions. 

\section{Experiments}
\subsection{Data}
 For this study we use publicly available data from HAM10000, a well curated data set of dermoscopy images collected specifically for use in the machine learning context~\cite{ham10000}. The full data set includes seven classes of skin lesions---in this study we concentrate on differentiating between benign naevus (moles) and melanoma, a particularly challenging clinical task. 
Our data set contains a total of 6017 images, with significant class imbalance: 5403 naevi and 614 malignant melanoma. We retain a balanced set of 200 images of each class as a hold out test set. It is worth noting that in the clinical context false negatives (predicting naevus when ground truth is melanoma) have far more serious consequences than false positives (predicting melanoma when ground truth is naevus). This means we need to ensure the class imbalance is addressed during training: a model trained and tested on the current distribution can achieve high accuracy (88\%) simply by always guessing naevus.
\subsection{Models}

The majority of publications in this area use transfer learning from Inception, pre-trained on Imagenet with an added pooling layer, dense layer and dropout---we follow suit for ease of comparison. We address class imbalance by first augmenting the melanoma images, obtaining a final set of 1656 melanoma images. We then sample 15 random subsets of 818 images from both classes and train a total of 30 models via a Bayesian hyper-parameter search---searching over learning rate, dropout, momentum, beta$_1$, beta$_2$, number of dense nodes, number of epochs, SGD and Adam\footnote{Details of augmentation, random data sampling, Bayesian hyper-parameter search, all code for training and experiments, including the final 30 trained models can be found here: https://github.com/KyleYoung1997/DNNorDermatologist}. The aim is to survey the landscape of possible models, giving us a selection of multiple networks to compare and explore rather than a single, cherry-picked one. The mean AUC over the 30 models is 85\% with a variance of 1.8\% and a mean recall of 87\%---a performance comparable to other published models in this context (e.g the model of~\cite{haenssle2018man} achieved AUC of 86\%)\footnote{Note that differences in test set size and distribution mean that direct comparison of model performance via AUC is of limited merit. However, as AUC is the standard metric reported in the literature, we include it here. Further comment can be found in the conclusions.}. Reported AUC for melanoma identification from dermoscopy images for dermatologists is around 79\%~\cite{haenssle2018man} and for primary care physicians even lower~\cite{raasch1999suspicious}. These results signal the fact that this is indeed a difficult task for which CNN decision support may prove useful.

\begin{figure}[htp]
\centering
\includegraphics[width=.25\textwidth]{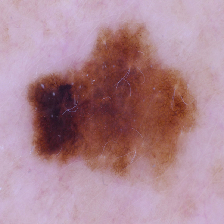}\hfill
\includegraphics[width=.25\textwidth]{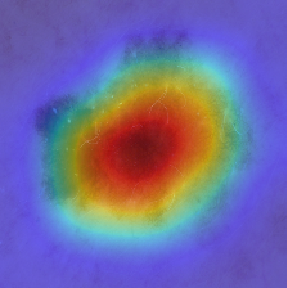}\hfill
\includegraphics[width=.25\textwidth]{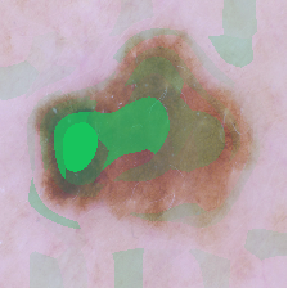}\hfill
\caption{Naevus mis-classified by all 30 models as a melanoma, with GradCAM and kernel SHAP saliency maps. Note there is more than one type of network within the lesion, a feature which can be a marker for melanoma. The GradCAM map (centre image) highlights a key deficiency of the method in this context: almost all of the lesion is obscured by the saliency map, rendering the ``explanation" clinically meaningless.}
\label{misclass}
\end{figure}

It is interesting to note that the variance across model accuracy (AUC) over the 30 models is relatively small at $1.8\%$. While these models share the same basic architecture, they have been trained on different sub-samples of the data using different hyper-parameters---thus are likely converging on different local optima. This is evidenced by the differences in mis-classified test images across the different models. Interestingly, seventeen images were consistently mis-classified: at least 25/30 models got the class label wrong. For example, the naevus in Fig.~\ref{misclass} was mis-classified by all 30 models as melanoma. Interestingly, this lesion does arguably satisfy one of the clinical criteria for melanoma.  A small human evaluation trial by 3 primary care physicians suggests these images are challenging: scores were 4/17, 5/17 and 6/17.

\subsection{GradCAM and Kernel SHAP}
GradCAM~\cite{gradcam2017} and Kernel SHAP~\cite{lundberg2017unified} are both model agnostic, local interpretability methods. While both highlight pixels that the trained network deems relevant for the final classification, they work in very different ways. GradCAM computes the gradient of the class-score (logit) with respect to the feature map of the final convolutional layer. Formally, consider each input image as a vector $x \in R_d$ where our model is a function $S: R_d \rightarrow R_c$, with $C$ the total number of classes. GradCAM provides an ``interpretability" map $I : R_d \rightarrow R_d$ that maps inputs to objects of the same dimension. If $A_k$ are feature maps obtained from the last convolutional layer, global average pooling of the gradients gives us a set of neuron importance maps $\alpha^{k}_c = \frac{1}{Z}\sum_i\sum_j \frac{\partial S}{\partial A^{k}_{ij}}$ and the final mask corresponds to a ReLU applied to a weighted linear combination of the feature maps and the importance maps: $I(x) =$ ReLU$(\sum_k\alpha^{k}_{c}A^k)$. More details and examples can be found in~\cite{gradcam2017}.

While there are a large variety of methods for applying saliency maps, recent work has shown that many are in fact independent of both the model weights and/or the class labels~\cite{adebayo2018sanity}. In these cases it is likely the model architecture itself is constraining the saliency maps to look falsely meaningful: frequently the maps just act as a variant of edge detector~\cite{adebayo2018sanity}. This is particularly dangerous in the context of skin cancer detection as features at the borders of lesions are often considered diagnostic for melanoma: saliency maps that highlight the edges of a lesion may be misconstrued as clinically meaningful. We use GradCAM in our analysis because it was one of the few methods that passed the recommended sanity checks (we also perform our own to double-check this particular context).

We also investigate the use of Kernel SHAP~\cite{lundberg2017unified}, an interpretability method that was not among those investigated in~\cite{adebayo2018sanity}, but has strong theoretical justification~\cite{Molnar2019}. A stronger agreement was found between Shapley explanations and human explanations when compared to two alternative popular saliency methods, LIME and DeepLIFT~\cite{lundberg2017unified}, further confirming that this is an appropriate method to explore. Based on Shapley values from co-operative game theory~\cite{shapley1953value}, the method assigns a fair attribution value to each feature based on the contribution that feature makes to the total prediction. The method is proven to be the unique mapping that satisfies a number of reasonable criteria and is calculated by considering interactions between all possible subsets of features. For $d$ features, calculating the Shapley value for a given feature $k$ will need to account for all $2^{d-1}$ subsets containing $k$. Thus a downside to the original approach is that it scales exponentially in the number of features. Consequently, we use an approximate, computationally feasible method: Kernel SHAP~\cite{lundberg2017unified}.

\begin{figure}[htp]
  \centering
    \includegraphics[scale = 0.36]{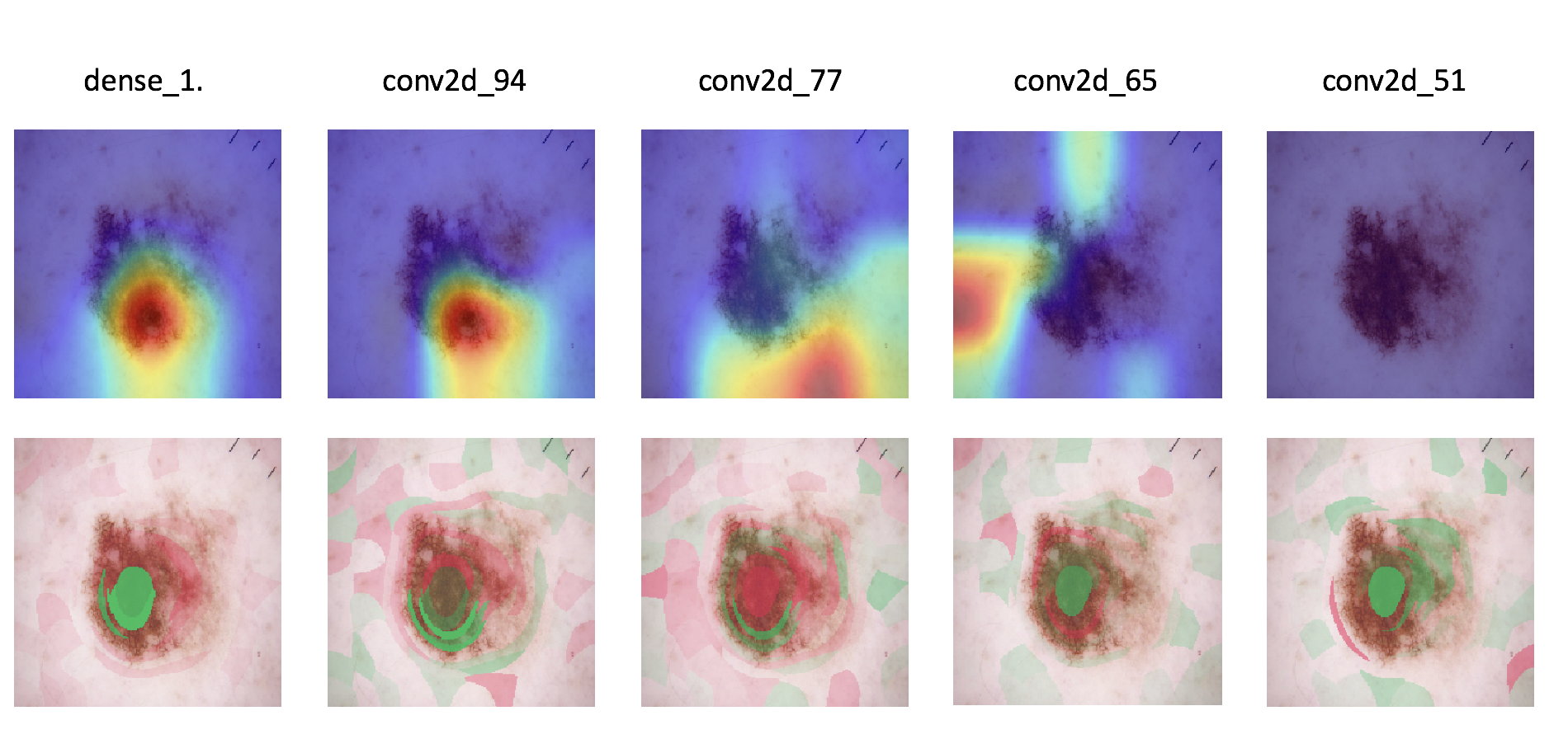} 
    \caption{Explanations following randomization of selected layers in the model. Changes demonstrate dependence of explanation on model weights. SSIM scores averaged over all images for GradCAM degraded across layers by 23\%, 4\%, 3\%, 2\%, 4\%. Differences were also seen for kernel SHAP: 17\%, 3\%, 5\%, 3\%, 7\%. Green signifies areas of positive contribution to a diagnosis of melanoma, red signifies negative.}
\label{modeldep}
\end{figure}
 \vspace{-9pt}
\subsection{Sanity checks}
We perform three simple sanity checks on GradCAM and Kernel SHAP to explore their performance in this context. (i) \textbf{Reproducibility:} we run the algorithms twice using the same randomly selected model and the same image, then compare images visually and using SSIM\footnote{Details on SSIM (Structural Similarity Index) can be found in~\cite{wang2004image}}. GradCAM saliency maps were unsurprisingly visually identical, with a perfect SSIM of 1, reflecting the deterministic nature of this algorithm. Kernel SHAP images were visually close to identical, but with SSIM less than perfect (mean 0.92, standard deviation 0.028). This small deviance is unsurprising given the method requires approximation via random sampling of subsets of features. (ii) \textbf{Model dependence:} using techniques inspired by~\cite{adebayo2018sanity} we randomize the weights of selective, progressively shallower layers in a randomly chosen model and recompute the GradCAM and Kernel SHAP images. The idea is to ensure that the saliency maps are not in fact independent of model weights. Visual comparison and SSIM scores verify that the maps are indeed model dependent, an example can be seen in Fig.~\ref{modeldep}. (iii) \textbf{Sensitivity:} we compare saliency maps from three models with the same AUC. This test serves to determine the sensitivity of the maps to model weights and also provides insight into differences across models of similar performance. Visual inspection shows variation across three models with identical AUCs of 85\% for both methods, with average variation in SSIM of 20\% for both GradCShrapenlAM and kernel SHAP. An example can be seen in Fig.~\ref{sens1}.

\subsection{Spurious correlation}
\begin{figure}[htp]
\centering
\includegraphics[width=.22\textwidth]{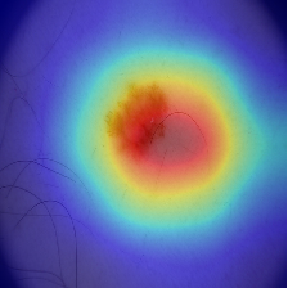}\hfill
\includegraphics[width=.22\textwidth]{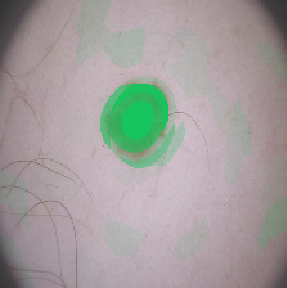}\hfill
\includegraphics[width=.22\textwidth]{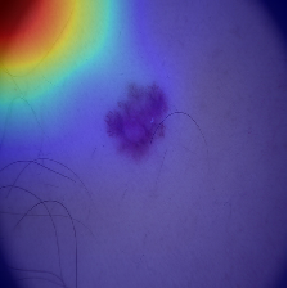}\hfill
\includegraphics[width=.22\textwidth]{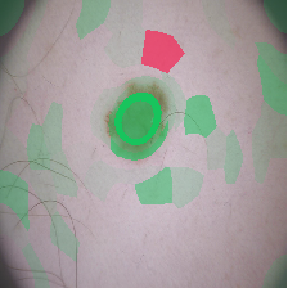}\hfill
\caption{GradCAM and kernel SHAP from two models with AUC 85\%. Model 1 correctly predicted melanoma with 0.999 confidence (first two images). Model 2 incorrectly predicted naevus with 0.996 confidence (second two images). The saliency maps indicate model 2 has learned to weight the class decision using a spurious correlation: the dark corners of the image.}
\label{sens1}
\end{figure}

 The saliency maps show that at this resolution the majority of images do not unambiguously capture clinically meaningful information. However, several images suggest that the model is indeed weighting the classification decision using spurious correlations. Notable examples include those images that highlight the dark corners of the images (e.g. Fig~\ref{sens1}).
 

\vspace{-9pt}

\subsection{Limitations} 

There are a number of limitations of this study. The small data size makes over-fitting more likely, thus increasing the chances that we would uncover spurious correlations. Additionally, while our small data size made many tasks computationally and practically feasible, for large test data sets this will not be the case---visual inspection to screen for spurious correlations will likely become impractical. An alternative approach would be to use these methods to provide feedback at the time of prediction: while a saliency map located on the lesion can not yet be viewed as justification that clinically meaningful correlations have been learned, a map that is clearly located on a clinically irrelevant region could be used to signal a prediction that should be ignored. 
Our study was also limited to models of a particular architecture, while we justify this as providing a point of comparison with existing published research, future work could include model architecture as a search hyper-parameter.

While the accuracy of our models is good and comparable to human accuracy, it is likely ensembled methods will improve accuracy further. It is difficult to envision how these interpretability methods could be applied meaningfully in this context. One alternative could be to use the maps themselves to regularize each of the models during training---methods such as GradMask suggest this may be possible~\cite{simpson2019gradmask}. Finally, there exists recent, alternative methods for implementing Shapley analysis that may well produce better results and permit the use of higher resolution images~\cite{chen2018shapley,aas2019explaining} . These experiments we leave for the future.

\section{Conclusions}
There is a significant literature comparing the performance of DNNs and dermatologists on test sets of dermoscopy images. These studies provide credibility for pursuing research in this area and the next task is to develop techniques that enable DNNs to become valued clinical decision support tools. We have shown that GradCAM and kernel SHAP maps pass some basic sanity checks and can provide insight into potential sources of bias. However, it is clear that more work is needed before these maps can provide clinically meaningful information. We have also shown that evaluating models according to AUC alone provides limited insight into the true nature of the performance of the model: saliency maps show that models with the same AUC can make predictions using completely different rationales.  

\section{Acknowledgements}
This work was supported by an Australian Research Council Centre of Excellence for Quantum Engineered Systems grant (CE 110001013).

%

%
%
%
\bibliographystyle{splncs04}
\bibliography{references.bib}

\end{document}